\documentclass[10pt,twocolumn,letterpaper]{article}
\usepackage{xcolor,colortbl}
\usepackage{cvpr}
\usepackage{times}
\usepackage{epsfig}
\usepackage{graphicx}
\usepackage{amsmath}
\usepackage{amssymb}
\usepackage{grffile}
\usepackage{enumitem}
\usepackage[hang]{footmisc}
\usepackage{placeins}
\usepackage{tikz}
\usepackage[pagebackref=true,breaklinks=true,letterpaper=true,colorlinks,citecolor=blue,bookmarks=false]{hyperref}
\usepackage{nopageno}

\definecolor{attr}{rgb}{0.85,0.88,0.90}
\definecolor{code}{rgb}{0.95,0.93,0.90}
\definecolor{Gray}{gray}{0.95}
\definecolor{LightCyan}{rgb}{0.88,1,1}
\newcolumntype{a}{>{\columncolor{Gray}}c}
\newcolumntype{b}{>{\columncolor{white}}c}

\newcommand{\dcnn}{FV-CNN\xspace}
\newcommand{\bcnn}{B-CNN\xspace}

\makeatletter
\g@addto@macro\normalsize{%
  \setlength\abovedisplayskip{0.7em}
  \setlength\belowdisplayskip{0.7em}
  \setlength\abovedisplayshortskip{0.7em}
  \setlength\belowdisplayshortskip{0.7em}
}
\makeatother

\tikzset{
 image label/.style={
   fill=white,
   text=black,
   font=\footnotesize,
   anchor=south east,
   xshift=-0.02cm,
   yshift=0.02cm,
   at={(0,0)}
 }
}

\newcommand{\puti}[2]
{%
\begin{tikzpicture}
\node[anchor=south east,inner sep=0] at (0,0) {#2};
\node[image label]{#1};
\end{tikzpicture}%
}

\cvprfinalcopy 
\def\cvprPaperID{2143} 

\ifcvprfinal\fi

\begin{document}

\title{Visualizing and Understanding Deep Texture Representations}
\author{
Tsung-Yu Lin\\
University of Massachusetts, Amherst \\
\small{\url{tsungyulin@cs.umass.edu}}
\and 
Subhransu Maji\\
University of Massachusetts, Amherst \\
\small{\url{smaji@cs.umass.edu}}
}
\maketitle
\begin{abstract}
A number of recent approaches have used deep convolutional neural networks (CNNs) to build texture representations. Nevertheless, it is still unclear how these models represent texture and invariances to categorical variations. This work conducts a systematic evaluation of recent CNN-based texture descriptors for recognition and attempts to understand the nature of invariances captured by these representations. First we show that the recently proposed bilinear CNN model~\cite{lin2015bilinear} is an excellent general-purpose texture descriptor and compares favorably to other CNN-based descriptors on various texture and scene recognition benchmarks. The model is translationally invariant and obtains better accuracy on the ImageNet dataset without requiring spatial jittering of data compared to corresponding models trained with spatial jittering. Based on recent work~\cite{gatys2015texture,mahendran2014understanding} we propose a technique to visualize pre-images, providing a means for understanding categorical properties that are captured by these representations. Finally, we show preliminary results on how a unified parametric model of texture analysis and synthesis can be used for attribute-based image manipulation, \eg to make an image more swirly, honeycombed, or knitted. The source code and additional visualizations are available at \url{http://vis-www.cs.umass.edu/texture}.
\end{abstract}

\section{Introduction}
The study of texture has inspired many of the early representations of images. The idea of representing texture using the statistics of image patches have led to the development of ``textons"~\cite{julesz83textons,leung01representing}, the popular ``bag-of-words" models~\cite{csurka04visual} and their variants such as the Fisher vector~\cite{perronnin07fisher} and VLAD~\cite{jegou10aggregating}. These fell out of favor when the latest generation of deep Convolutional Neural Networks~(CNNs) showed significant improvements in recognition performance over a wide range of visual tasks~\cite{chatfield14return,girshick14rich,jia2014caffe,razavin14cnn-features}. Recently however, the interest in texture descriptors have been revived by architectures that combine aspects of texture representations with CNNs. For instance, Cimpoi \etal~\cite{cimpoi2016deep} showed that Fisher vectors built on top of CNN activations lead to better accuracy and improved domain adaptation not only for texture recognition, but also for scene categorization, object classification, and fine-grained recognition.

\begin{figure}
\begin{center}
\renewcommand{\arraystretch}{0.8}
\setlength{\tabcolsep}{1pt}
\begin{tabular}{ccc}
\puti{dotted}{\includegraphics[width=0.32\linewidth]{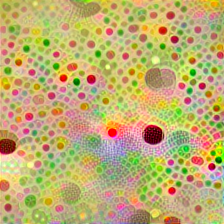}} & 
\puti{water}{\includegraphics[width=0.32\linewidth]{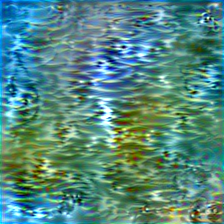}} &
\puti{landromat}{\includegraphics[width=0.32\linewidth]{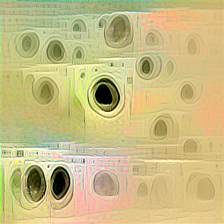}} \\ 
\puti{honeycombed}{\includegraphics[width=0.32\linewidth]{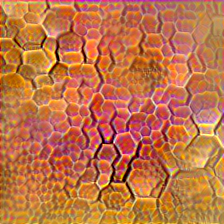}} & 
\puti{wood}{\includegraphics[width=0.32\linewidth]{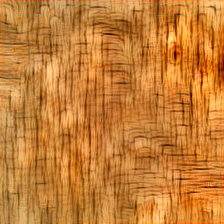}} &
\puti{bookstore}{\includegraphics[width=0.32\linewidth]{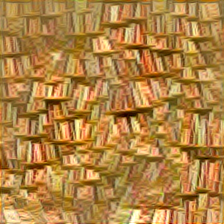}} \\ 
\end{tabular}
\end{center}
\caption{\label{fig:inverse} \textbf{How is texture represented in deep models?} Visualizing various categories by inverting the bilinear CNN model~\cite{lin2015bilinear} trained on DTD~\cite{cimpoi14describing}, FMD~\cite{sharan09material}, and MIT Indoor dataset~\cite{quattoni09recognizing} (each column from left to right). These images were obtained by starting from a random image and adjusting it though gradient descent to obtain high log-likelihood for the given category label using a multi-layer bilinear CNN model (See Sect.~\ref{sec:methodology} for details). \emph{Best viewed in color and with zoom.}}
\end{figure}

Despite their success little is known how these models represent \emph{invariances} at the image and category level. Recently, several attempts have been made in order to understand CNNs by visualizing the layers of a trained model~\cite{zeiler2014visualizing}, studying the invariances by inverting the model~\cite{dosovitskiy2015inverting,mahendran2014understanding,simonyan2013deep}, and evaluating the performance of CNNs for various recognition tasks. \emph{In this work we attempt to provide a similar understanding for CNN-based texture representations}. Our starting point is the bilinear CNN model of our previous work~\cite{lin2015bilinear}. The technique builds an orderless image representation by taking the location-wise outer product of image features extracted from CNNs and aggregating them by averaging. The model is closely related to Fisher vectors but has the advantage that gradients of the model can be easily computed allowing fine-tuning and inversion. Moreover, when the two CNNs are identical the bilinear features capture correlations between filter channels, similar to the early work in parametric texture representation of Portilla and Simoncelli~\cite{portilla2000parametric}.

Our \emph{first contribution} is a systematic study of bilinear CNN features for texture recognition. Using the \emph{Flickr Material Dataset} (FMD)~\cite{sharan09material}, \emph{Describable Texture Dataset} (DTD)~\cite{cimpoi14describing} and \emph{KTH-T2b}~\cite{caputo05class} we show that it performs favorably to Fisher vector CNN model~\cite{cimpoi2016deep}, which is the current state of the art. Similar results are also reported for scene classification on the \emph{MIT indoor scene} dataset~\cite{quattoni09recognizing}. We also evaluate the role of different layers, effect of scale, and fine-tuning for texture recognition. Our experiments reveal that multi-scale information helps, but also that features from different layers are complementary and can be combined to improve performance in agreement with recent work~\cite{bharath2015hypercolumns,long2014fully,mostajabi2014feedforward}.

Our \emph{second contribution} is to investigate the role of translational invariance of these models due to the orderless pooling of local descriptors. Recently, we showed~\cite{lin2015bilinear} that bilinear CNNs initialized by pre-training a standard CNN (\eg, VGG-M~\cite{chatfield14return}) on ImageNet, truncating the network at a lower convolutional layer (\eg, \emph{conv5}), adding bilinear pooling modules, followed by domain-specific fine-tuning leads to significant improvements in accuracy for a number of fine-grained recognition tasks. These models capture localized feature interactions in a translationally invariant manner which is useful for making fine-grained distinctions between species of birds or models of cars. However, it remains unclear what the tradeoffs are between explicit translational invariance in these models versus implicit invariance obtained by spatial jittering of data during training. To this end we conduct experiments on the ImageNet LSVRC 2012 dataset~\cite{deng09imagenet} by training several models using different amounts of data augmentation. Our experiments reveal that bilinear CNN models can be trained from \emph{scratch}, resulting in better accuracy \emph{without} requiring spatial jittering of data than the corresponding CNN architectures that consist of standard ``fully-connected" layers trained with jittering.

Our \emph{third contribution} is a technique to ``invert" these models to construct invariant inputs and visualize pre-images for categories. Fig.~\ref{fig:inverse} shows the inverse images for various categories -- \emph{materials} such as wood and water, \emph{describable attributes} such as honeycombed and dotted, and \emph{scene categories} such as laundromat and bookstore. These images reveal what categorical properties are learned by these texture models. Recently, Gatys \etal~\cite{gatys2015texture} showed that bilinear features (they call it the Gram matrix representation) extracted from various layers of the ``verydeep VGG network"~\cite{simonyan14very} can be inverted for texture synthesis. The synthesized results are visually appealing, demonstrating that the convolutional layers of a CNN capture textural properties significantly better than the first and second order statistics of wavelet coefficients of Portilla and Simoncelli. However, the approach remains impractical since it requires hundreds of CNN evaluations and is orders of magnitude slower than non-parametric patch-based methods such as image quilting~\cite{Efros01}. We show that the two approaches are complementary and one can significantly speed up the convergence of gradient-based inversion by initializing the inverse image using image quilting. The global adjustment of the image through gradient descent also removes many artifacts that quilting introduces. 

Finally, we show a novel application of our approach for \emph{image manipulation} with \emph{texture attributes}. A unified parametric model of texture representation and recognition allows us to adjust an image with high-level attributes -- to make an image more \emph{swirly} or \emph{honeycombed}, or generate hybrid images that are a combination of multiple texture attributes, \eg, \emph{chequered} and  \emph{interlaced}.

\subsection{Related work}

Texture recognition is a widely studied area. Current state-of-the-art results on texture and material recognition are obtained by hybrid approaches that build orderless representations on top of CNN activations. Cimpoi \etal~\cite{cimpoi2016deep} use Fisher vector pooling for material and scene recognition, Gong \etal~\cite{gong14multi-scale} use VLAD pooling for scene recognition, \etc. Our previous work~\cite{lin2015bilinear} proposed a general orderless pooling architecture called the \emph{bilinear CNN} that outperforms Fisher vector on many fine-grained datasets. These descriptors are inspired by early work on texture representations~\cite{csurka04visual,leung01representing,perronnin07fisher,jegou10aggregating} that were built on top of wavelet coefficients, linear filter bank responses, or SIFT features~\cite{lowe99object}.

Texture synthesis has received attention from both the vision and graphics communities due to its  numerous applications. Heeger and Bergen~\cite{DBLP:conf/siggraph/HeegerB95} synthesized texture images by histogram matching. Portilla and Simoncelli were one of the early proponents of parametric approaches. The idea is to represent texture as the first and second order statistics of various filter bank responses (\eg, wavelets, steerable pyramids, \etc). However, these approaches were outperformed by simpler non-parametric approaches. For instance, Efros and Lueng~\cite{efros1999texture} proposed a pixel-by-pixel synthesis approach based on sampling similar patches -- the method was simple and effective for a wide range of textures. Later, Efros and Freeman proposed a quilting-based approach that was significantly faster~\cite{Efros01}. A number of other non-parametric approaches have been proposed for this problem~\cite{kwatra2003graphcut,wei2000fast}. Recently, Gatys \etal showed that replacing the linear filterbanks by CNN filterbanks results in better reconstructions. Notably, the Gram matrix representation used in their approach is identical to the bilinear CNN features of Lin \etal, suggesting that these features might be good for texture recognition as well. However for synthesis, the parametric approaches remain impractical as they are orders of magnitude slower than non-parametric approaches.

Understanding CNNs through visualizations has also been widely studied given their remarkable performance. Zieler and Fergus~\cite{zeiler2014visualizing} visualize CNNs using the top activations of filters and show per-pixel heatmaps by tracking the position of the highest responses. Simonyan and Zisserman~\cite{simonyan2013deep} visualize parts of the image that cause the highest change in class labels computed by back-propagation. Mahendran and Vedaldi~\cite{mahendran2014understanding} extend this approach by introducing natural image priors which result in inverse images that have fewer artifacts. Dosovitskiy and Brox~\cite{dosovitskiy2015inverting} propose a  ``deconvolutional network" to invert a CNN in a feed-forward manner. However, the approach tends to produce blurry images since the inverse is not uniquely defined.

Our approach is also related to prior work on editing images based on example images. Ideas from patch-based texture synthesis have been extended in a number of ways to modify the style of the image based on an example~\cite{hertzmann2001image}, adjusting texture synthesis based on the content of another image~\cite{criminisi2004region,Efros01}, \etc. Recently, in a separate work, Gatys \etal~\cite{gatys2015neural} proposed a ``neural style" approach that combines ideas from inverting CNNs with their work on texture synthesis. They generate images that match the style and content of two different images producing compelling results. Although the approach is not practical compared to existing patch-based methods for editing styles, it provides a basis for a rich parametric model of texture. We describe an novel approach to manipulate images with high-level attributes and show several examples of editing images with texture attributes. There is relatively little prior work on manipulating the content of an image using semantic attributes. 

\section{Methodology and overview}\label{sec:methodology}
We describe our framework for parametric texture recognition, synthesis, inversion, and attribute-based manipulation using CNNs. For an image ${\cal I}$ one can compute the activations of the CNN at a given layer $r_i$ to obtain a set of features $F_{r_i} = \{f_j\}$ indexed by their location $j$. The bilinear feature $B_{r_i}({\cal I})$ of  $F_{r_i}$ is obtained by computing the outer product of each feature $f_j$ with itself and aggregating them across locations by averaging, \ie, 
\begin{equation}
B_{r_i}({\cal I}) = \frac{1}{N} \sum_{j=1}^N f_j f_j^T.
\end{equation}

The bilinear feature (or the Gram matrix representation) is an \emph{orderless representation} of the input image and hence is suitable for modeling texture. Let $r_i,i=1,\ldots, n,$ be the index of the $i^{th}$ layer with the bilinear feature representation $B_{r_i}$. Gatys \etal~\cite{gatys2015texture} propose a method for texture synthesis from an input image ${\cal I}$ by obtaining an image $\mathbf{x} \in \mathbb{R}^{H\times W \times C} $ that matches the bilinear features at various layers by solving the following optimization:

\begin{equation}\label{obj:tex}
\min_\mathbf{x} \sum_{i=1}^n \alpha_i L_1\left( B_{r_i},\hat{B}_{r_i}\right)+ \gamma \Gamma(\mathbf{x}).
\end{equation}

Here, $\hat{B}_{r_i} = B_{r_i}({\cal I})$, $\alpha_i$ is the weight of the $i^{th}$ layer, $\Gamma(\mathbf{x})$ is a natural image prior such as the total variation norm (TV norm), and $\gamma$ is the weight on the prior. Note that we have dropped the implicit dependence of $B_{r_i}$ on $\mathbf{x}$ for brevity. Using the squared loss-function $L_1(x,y) = \sum (x_i-y_i)^2$ and starting from a random image where each pixel initialized with a \emph{i.i.d} zero mean Gaussian, a local optimum is reached through gradient descent. The authors employ L-BFGS, but any other optimization method can be used (\eg, Mahendran and Vedaldi~\cite{mahendran2014understanding} use stochastic gradient descent with momentum). 

Prior work on minimizing the reconstruction error with respect to the ``un-pooled" features $F_{r_i}$ has shown that the \emph{content} of the image in terms of the color and spatial structure is well-preserved even in the higher convolutional layers. Recently, Gatys \etal in a separate work~\cite{gatys2015neural} synthesize images that match the style and content of two different images ${\cal I}$ and ${\cal I'}$ respectively by minimizing a weighted sum of the texture and content reconstruction errors: 

\begin{equation}\label{obj:tex}
\min_\mathbf{x} \lambda L_1 \left( F_{s},\hat{F}_{s}\right) + \sum_{i=1}^n \alpha_i L_1\left( B_{r_i},\hat{B}_{r_i}\right)+ \gamma \Gamma(\mathbf{x}).
\end{equation}

Here $\hat{F}_s = F_s({\cal I'})$ are features from a layer $s$ from which the target content features are computed for an image ${\cal I'}$. 

The bilinear features can also be used for predicting attributes by first normalizing the features (signed square-root and $\ell_2$) and training a linear classifier in a supervised manner~\cite{lin2015bilinear}. Let $l_i : i=1,\ldots, m$ be the index of the $i^{th}$ layer from which we obtain attribute prediction probabilities $C_{l_i}$. The prediction layers may be different from those used for texture synthesis. Given a target attribute $\hat{C}$ we can obtain an image that matches the target label and is similar to the texture of a given image ${\cal I}$ by solving the following optimization:

\begin{equation}
\min_\mathbf{x} \sum_{i=1}^n \alpha_i L_1\left( B_{r_i},\hat{B}_{r_i}\right) + \beta \sum_{i=1}^{m} L_2\left( C_{l_i}, \hat{C}\right) + \gamma \Gamma(\mathbf{x}).
\end{equation}

Here, $L_2$ is a loss function such as the \emph{negative log-likelihood} of the label $\hat{C}$ and $\beta$ is a tradeoff parameter. If multiple targets $\hat{C}_j$ are available then the losses can be blended with weights $\beta_j$ resulting in the following optimization:
\begin{equation}\label{eq:obj}
\min_\mathbf{x} \sum_{i=1}^n \alpha_i L_1\left( B_{r_i},\hat{B}_{r_i}\right) + \beta_j \sum_{i=1, j}^{m} L_2\left( C_{l_i}, \hat{C}_j\right) + \gamma \Gamma(\mathbf{x}).
\end{equation}

\setlength\footnotemargin{5pt}
\paragraph{Implementation details.} We use the 16-layer VGG network~\cite{simonyan14very} trained on ImageNet for all our experiments. For the image prior $\Gamma(\mathbf{x})$ we use the $\text{TV}_\beta$ norm with $\beta=2$: 
\begin{equation}
   \Gamma(\mathbf{x}) = \sum_{i,j} \left((x_{i,j+1} - x_{i,j})^2 + (x_{i+1,j} - x_{i,j})^2\right)^{\frac{\beta}{2}}.
\end{equation}
The exponent $\beta=2$ was empirically found to lead to better reconstructions in ~\cite{mahendran2014understanding} as it leads to fewer ``spike" artifacts than $\beta=1$. In all our experiments, given an input image we resize it to 224$\times$224 pixels before computing the target bilinear features and solve for $\mathbf{x} \in \mathbb{R}^{224\times 224\times 3}$. This is primarily for speed since the size of the bilinear features are independent of the size of the image. Hence, an output of any size can be obtained by minimizing Eqn.~\ref{eq:obj}. We use L-BFGS for optimization and compute the gradients of the objective with respect to $\mathbf{x}$ using back-propagation. One detail we found to be critical for good reconstructions is that we $\ell_1$ normalize the gradients with respect to each of the $L_1$ loss terms to balance the losses during optimization. Mahendran and Vedaldi~\cite{mahendran2014understanding} suggest normalizing each $L_1$ loss term by the $\ell_2$ norm of the target feature $\hat{B}_{r_i}$. Without some from of  normalization the losses from different layers are of vastly different scales leading to numerical stability issues during optimization. 

Using this framework we: (i) study the effectiveness of bilinear features $B_{r_i}$ extracted from various layers of a network for texture and scene recognition (Sect.~\ref{sec:recognition}), (ii) investigate the nature of invariances of these features by evaluating the effect of training with different amounts of data augmentation (Sect.~\ref{sec:imagenet}), (iii) provide insights into the learned models by inverting them (Sect.~\ref{sec:understanding}), and (iv) show results for modifying the content of an image with texture attributes (Sect.~\ref{sec:manipulation}). We conclude in Sect.~\ref{s:conclusion}.

\newcolumntype{L}[1]{>{\raggedright\let\newline\\\arraybackslash\hspace{0pt}}m{#1}}
\newcolumntype{C}[1]{>{\centering\let\newline\\\arraybackslash\hspace{0pt}}m{#1}}
\newcolumntype{R}[1]{>{\raggedleft\let\newline\\\arraybackslash\hspace{0pt}}m{#1}}

\section{Texture recognition}\label{sec:recognition}

\noindent
In this section we evaluate the bilinear CNN (\bcnn) representation for texture recognition and scene recognition.

\paragraph{Datasets and evaluation.}\label{s:dataset}
We experiment on three texture datasets -- the \emph{Describable Texture Dataset}~(DTD)~\cite{cimpoi14describing}, \textit{Flickr Material Dataset} (FMD)~\cite{sharan09material}, and \emph{KTH-TISP2-b} (KTH-T2b)~\cite{caputo05class}. DTD consists of 5640 images labeled with 47 describable texture attributes.  FMD consists of 10 material categories, each of which contains 100 images. Unlike DTD and FMD where images are collected from the Internet, KTH-T2b contains 4752 images of 11 materials captured under controlled scale, pose, and illumination. The KTH-T2b dataset splits the images into four samples for each category. We follow the standard protocol by training on one sample and test on the remaining three. On DTD and FMD, we randomly divide the dataset into 10 splits and report the mean accuracy across splits. Besides these, we also evaluate our models on \textit{MIT indoor scene} dataset~\cite{quattoni09recognizing}. Indoor scenes are weakly structured and orderless texture representations have been shown to be effective here. The dataset consists of 67 indoor categories and a defined training and test split.

\paragraph{Descriptor details and training protocol.} Our features are based on the ``verydeep VGG network"~\cite{simonyan14very} consisting of 16 convolutional layers pre-trained on the ImageNet dataset. The FV-CNN builds a Fisher Vector representation by extracting CNN filterbank responses from a particular layer of the CNN using 64 Gaussian mixture components, identical to setup of Cimpoi \etal~\cite{cimpoi2016deep}. The B-CNN features are similarly built by taking the location-wise outer product of the filterbank responses and average pooling across all locations (identical to B-CNN [D,D] in Lin \etal~\cite{lin2015bilinear}). Both these features are passed through signed square-root and $\ell_2$ normalization which has been shown to improve performance. During training we learn one-vs-all SVMs (trained with SVM hyperparameter $C=1$) and weights scaled such that the median positive and negative class scores in the training data is $+1$ and $-1$ respectively. At test time we assign the class with the highest score. Our code in implemented using MatConvNet~\cite{vedaldi2015matconvnet} and VLFEAT~\cite{vedaldi2010vlfeat} libraries.

\subsection{Experiments}

\noindent
The following are the main conclusions of the experiments:

\paragraph{1.~B-CNN compares favorably to FV-CNN.}
Tab.~\ref{tab:results} shows results using the B-CNN and FV-CNN on various datasets. Across all scales of the input image the performance using B-CNN and FV-CNN is virtually identical. The FV-CNN multi-scale results reported here are comparable ($\pm1\%$) to the results reported in Cimpoi \etal~\cite{cimpoi2016deep} for all datasets except KTH-T2b ($-4\%$). These differences in results are likely due to the choice of the CNN \footnote{they use the \emph{conv5\_4} layer of the 19-layer VGG network.} and the range of scales. These results show that the bilinear pooling is \emph{as good as} the Fisher vector pooling for texture recognition. One drawback is that the FV-CNN features with 64 GMM components has half as many dimensions (64$\times$2$\times$256) as the bilinear features (256$\times$256). However, it is known that these features are highly redundant and their dimensionality can be reduced by an order of magnitude without loss in performance~\cite{gao2015compact,lin2015bilinear}.

\paragraph{2.~Multi-scale analysis improves performance.} Tab.~\ref{tab:results} shows the results by combining features from multiple scales $2^s, s\in$ \{1.5:-0.5:-3\} relative to the 224$\times$224 image. We discard scales for which the image is smaller than the size of the receptive fields of the filters, or larger than $1024^2$ pixels for efficiency. Multiple scales consistently lead to an improvement in accuracy.

\paragraph{3.~Higher layers perform better.} Tab.~\ref{tab:layers} shows the performance using various layers of the CNN. The accuracy improves using the higher layers in agreement with~\cite{cimpoi2016deep}. 

\paragraph{4.~Multi-layer features improve performance.} By combining features from all the layers we observe a small but significant improvement in accuracy on DTD $69.9\%$~$\rightarrow$~$70.7\%$ and on MIT indoor from $72.8\% \rightarrow  74.9\%$. This suggests that the features from multiple layers capture complementary information and can be combined to improve performance. This is in agreement with the ``hypercolumn" approach of Hariharan \etal~\cite{bharath2015hypercolumns}.

\paragraph{5.~Fine-tuning leads to a small improvement.} On the MIT indoor dataset fine-tuning the network using the B-CNN architecture leads to a small improvement $72.8\% \rightarrow 73.8\%$ using \emph{relu5\_3} and $s=1$. Fine-tuning on texture datasets led to insignificant improvements which might be attributed to their small size. On larger and specialized datasets, such as fine-grained recognition, the effect of fine-tuning can be significant~\cite{lin2015bilinear}. 

\begin{table}[t]
\small
\setlength{\tabcolsep}{4pt}
\begin{center}
\begin{tabular}{l|ccc|ccc}
 & \multicolumn{3}{c|}{\bf{\dcnn}} & \multicolumn{3}{c}{\bf{\bcnn}}  \\ 
\cline{2-7} 
\textbf{Dataset} & $s=1$ & $s=2$ & $ms$ & $s=1$ & $s=2$ & $ms$ \\ 
\hline 
DTD & $67.8$  & $70.6$ & $73.6$  & $69.6$ & $71.5$ & $72.9$\\ 
        & $^{\pm0.9}$   & $^{\pm0.9}$ & $^{\pm1.0}$  & $^{\pm0.7}$ & $^{\pm0.8}$ & $^{\pm0.8}$\\ 
FMD & $75.1$  & $79.0$  & $80.8$ & $77.8$ & $80.7$  & $81.6$ \\ 
 & $^{\pm2.3}$  & $^{\pm1.4}$  & $^{\pm1.7}$ & $^{\pm1.9}$ & $^{\pm1.5}$  & $^{\pm1.7}$ \\ 
KTH-T2b & $74.8$  & $75.9$ & $77.9$ & $75.1$ & $76.4$ & $77.9$ \\ 
 & $^{\pm2.6}$  & $^{\pm2.4}$ & $^{\pm2.0}$ & $^{\pm2.8}$ & $^{\pm3.5}$ & $^{\pm3.1}$ \\ 
MIT indoor& $70.1$ & $78.2$  & $78.5$  & $72.8$ & $77.6$  & $79.0$ \\ 
\end{tabular}
\end{center}
\caption{\label{tab:results}
\textbf{Comparison of B-CNN and FV-CNN.} We report mean per-class accuracy on DTD, FMD, KTH-T2b and MIT indoor datasets using FV-CNN and B-CNN representations constructed on top of \emph{relu5\_3} layer outputs of the 16-layer VGG network~\cite{simonyan14very}. Results are reported using input images at different scales: $s=1$, $s=2$ and $ms$ are with images resized to 224$\times$224, 448$\times$448 and pooled across multiple scales respectively. }
\end{table}

\begin{table}[!t]
\small
\begin{center}
\begin{tabular}{l|c|c|c|c}
\textbf{Dataset} & \textbf{\textit{relu2\_2}} & \textbf{\textit{relu3\_3}} & \textbf{\textit{relu4\_3}} & \textbf{\textit{relu5\_3}} \\
\hline
DTD & $42.9$   &  $59.0$ & $68.8$  & $69.9$ \\ 
FMD & $49.6$  &  $62.2$ & $73.4$ & $80.2$ \\ 
KTH-T2b & $59.9$  & $71.3$  & $78.8$ & $79.0$ \\ 
MIT indoor & $32.2$ & $54.5$ & $71.1$& $72.8$ \\
\end{tabular}
\end{center}
\caption{\label{tab:layers}
\textbf{Layer by layer performance.} The classification accuracy using \bcnn features based on the outputs of different layers on several datasets using input at $s=1$, i.e. 224$\times$224 pixels. The numbers are reported on the first split of all datasets.}
\end{table}

\section{The role of translational invariance}\label{sec:imagenet}

Earlier experiments on B-CNN and FV-CNN were reported using pre-trained networks. Here we experiment with training a B-CNN model \emph{from scratch} on the ImageNet LSRVC 2012 dataset. We experimenting with the effect of spatial jittering of training data on the classification performance. For these experiments we use the VGG-M~\cite{chatfield14return} architecture which performs better than AlexNet~\cite{krizhevsky12imagenet} with a moderate decrease in classification speed. For the B-CNN model we replace the last two fully-connected layers with a linear classifier and softmax layer on the outputs of the square-root and $\ell_2$ normalized bilinear features of the \emph{relu5} layer outputs. The evaluation speed for B-CNN is 80\% of that of the standard CNN, hence the overall training times for both architectures are comparable.

We train various models with different amounts of spatial jittering -- ``f1" for flip, ``f5" for flip + 5 translations and ``f25" for flip + 25 translations. In each case the training is done using stochastic sampling where one of the jittered copies is randomly selected for each example. The network parameters are randomly initialized and trained using stochastic gradient descent with momentum for a number of epochs. We start with a high learning rate and reduce it by a factor of 10 when the validation error stops decreasing. We stop training when the validation error stops decreasing.

\begin{figure}
\begin{center}
\includegraphics[width=1\linewidth]{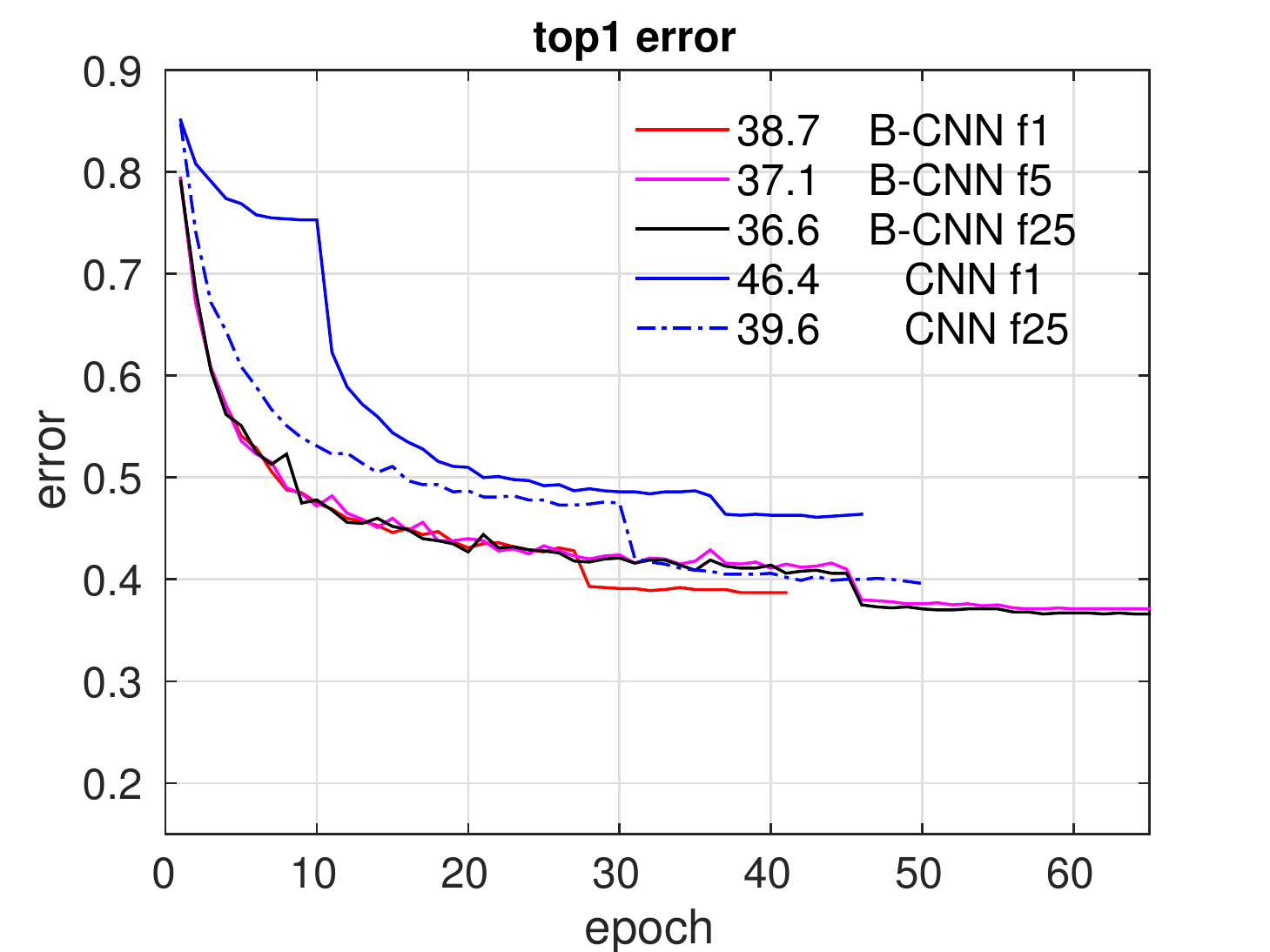}
\end{center}
\vspace{-0.1in}
\caption{\label{fig:imagenet} \textbf{Effect of spatial jittering on ImageNet LRVC 2012 classification.} The top1 validation error on a single center crop on ImageNet dataset using the VGG-M network and the corresponding \bcnn model. The networks are trained with different levels of data jittering: ``f1", ``f5", and ``f25" indicating flip, flip + 5 translations, and flip + 25 translations respectively.}
\end{figure}

Fig.~\ref{fig:imagenet} shows the ``top1" validation errors and compares the \bcnn network to the standard VGG-M model. The validation error is reported on a single center cropped image. Note that we train all networks with neither PCA color jittering nor batch normalization and our baseline results are within $2\%$ of the top1 errors reported in~\cite{chatfield14return}. The VGG-M model achieves $46.4\%$ top1 error with flip augmentation during training. The performance improves significantly to $39.6\%$ with f25 augmentation. As fully connected layers in a standard CNN network encode spatial information, the model loses performance without spatial jittering. For \bcnn network, the model achieves $38.7\%$ top1 error with f1 augmentation, outperforming VGG-M with f25 augmentation. With more augmentations, \bcnn model improves top1 error by $1.6\%$ ($38.7\%\rightarrow37.1\%$). Going from f5 to f25, \bcnn model improves marginally by $<1\%$. The results show that \bcnn feature is discriminative and robust to translation. With a small amount of data jittering, \bcnn network achieves fairly good performance, suggesting that explicit translation invariance might be preferable to the implicit invariance obtained by data jittering.

\section{Understanding texture representations}\label{sec:understanding}
In this section we aim to understand B-CNN texture representation by synthesizing \emph{invariant images}, i.e. images that are nearly identical to a given image according to the bilinear features, and \emph{inverse images} for a given category. 

\begin{figure}
\begin{center}
\renewcommand{\arraystretch}{0}
\setlength{\tabcolsep}{0pt}
\begin{tabular}{ccc}
\includegraphics[width=0.33\linewidth]{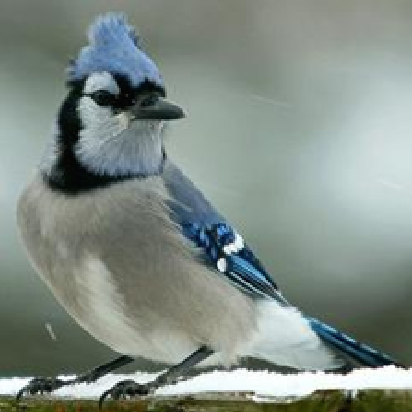} & 
\includegraphics[width=0.33\linewidth]{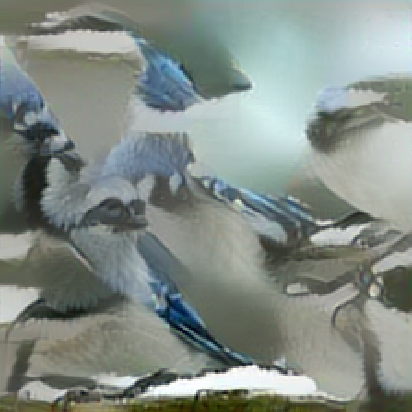} & 
\includegraphics[width=0.33\linewidth]{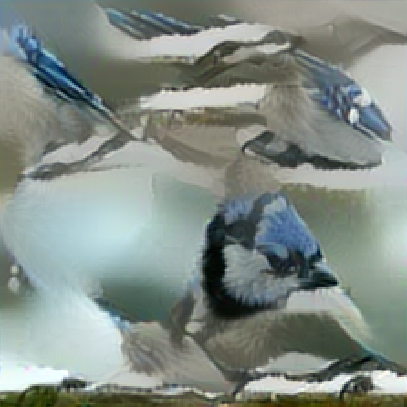} \\
\includegraphics[width=0.33\linewidth]{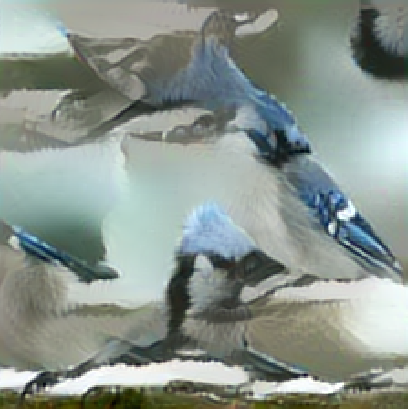} & 
\includegraphics[width=0.33\linewidth]{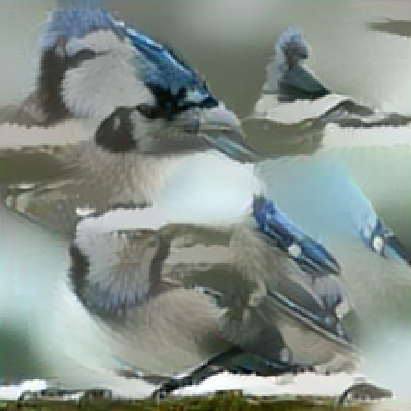} & 
\includegraphics[width=0.33\linewidth]{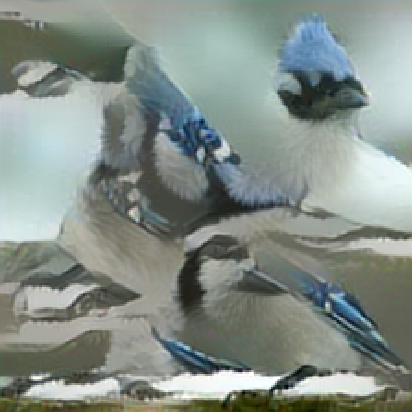} \\
\end{tabular}
\end{center}
\vspace{-0.1in}
\caption{\label{fig:invariance} \textbf{Invariant inputs.} These six images are virtually identical when compared using the bilinear features of layers \textit{relu1\_1, relu2\_1, relu3\_1, relu4\_1, relu5\_1} of the VGG network~\cite{simonyan14very}.}
\vspace{-0.1in}
\end{figure}

\paragraph{Visualizing invariant images for objects.} We use \textit{relu1\_1, relu2\_1, relu3\_1, relu4\_1, relu5\_1} layers for texture representation. Fig.~\ref{fig:invariance} shows several invariant images to the image on the top left, \ie these images are virtually identical as far as the bilinear features for these layers are concerned. Translational invariance manifests as shuffling of patches but important local structure is preserved within the images. These images were obtained using $\gamma=1e-6$ and $\alpha_i=1~\forall i$ in Eqn.~\ref{eq:obj}. We found that as long as some higher and lower layers are used together the synthesized textures look reasonable, similar to the observations of Gatys \etal.

\begin{figure*}
\begin{center}
\renewcommand{\arraystretch}{0.8}
\begin{tabular}{cc}
\setlength{\tabcolsep}{1pt}
\includegraphics[width=0.61\linewidth]{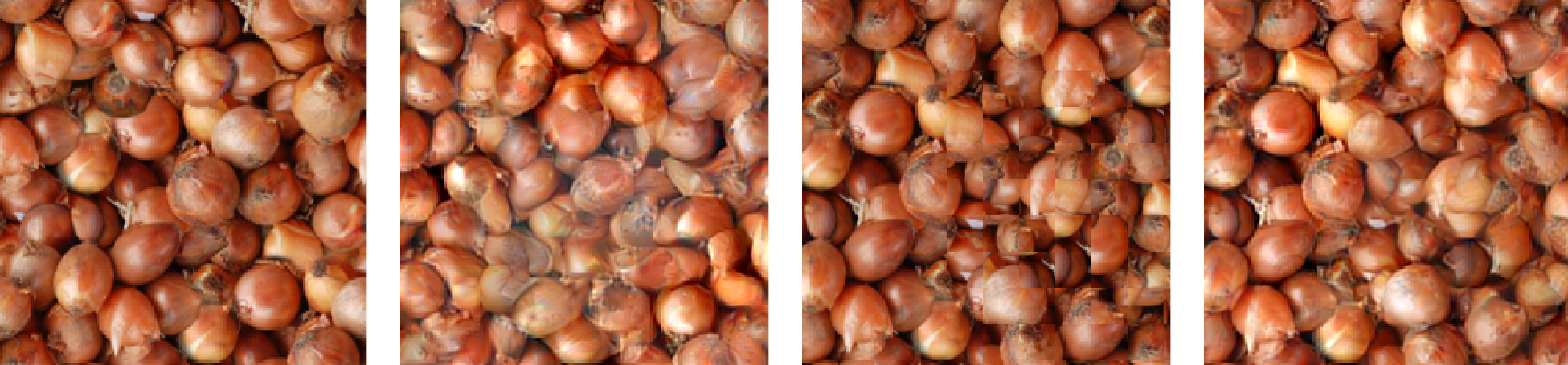} & 
\includegraphics[width=0.2\linewidth]{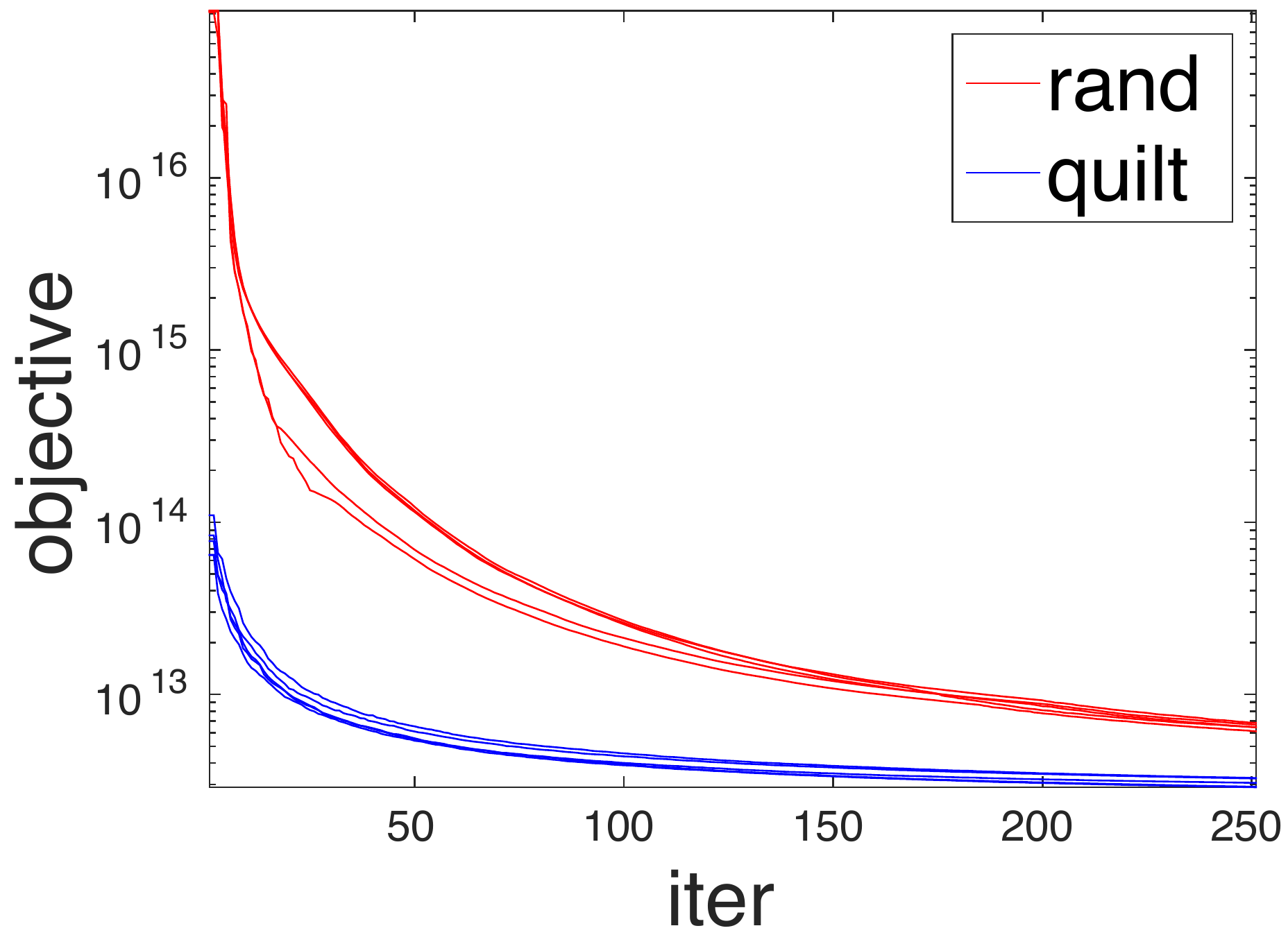} \\
\includegraphics[width=0.61\linewidth]{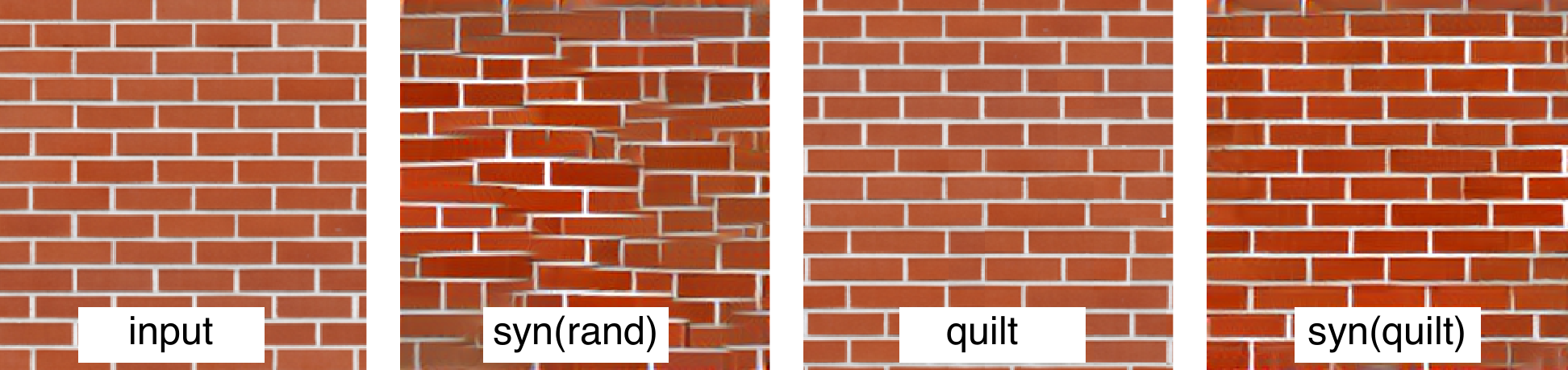} &
\includegraphics[width=0.2\linewidth]{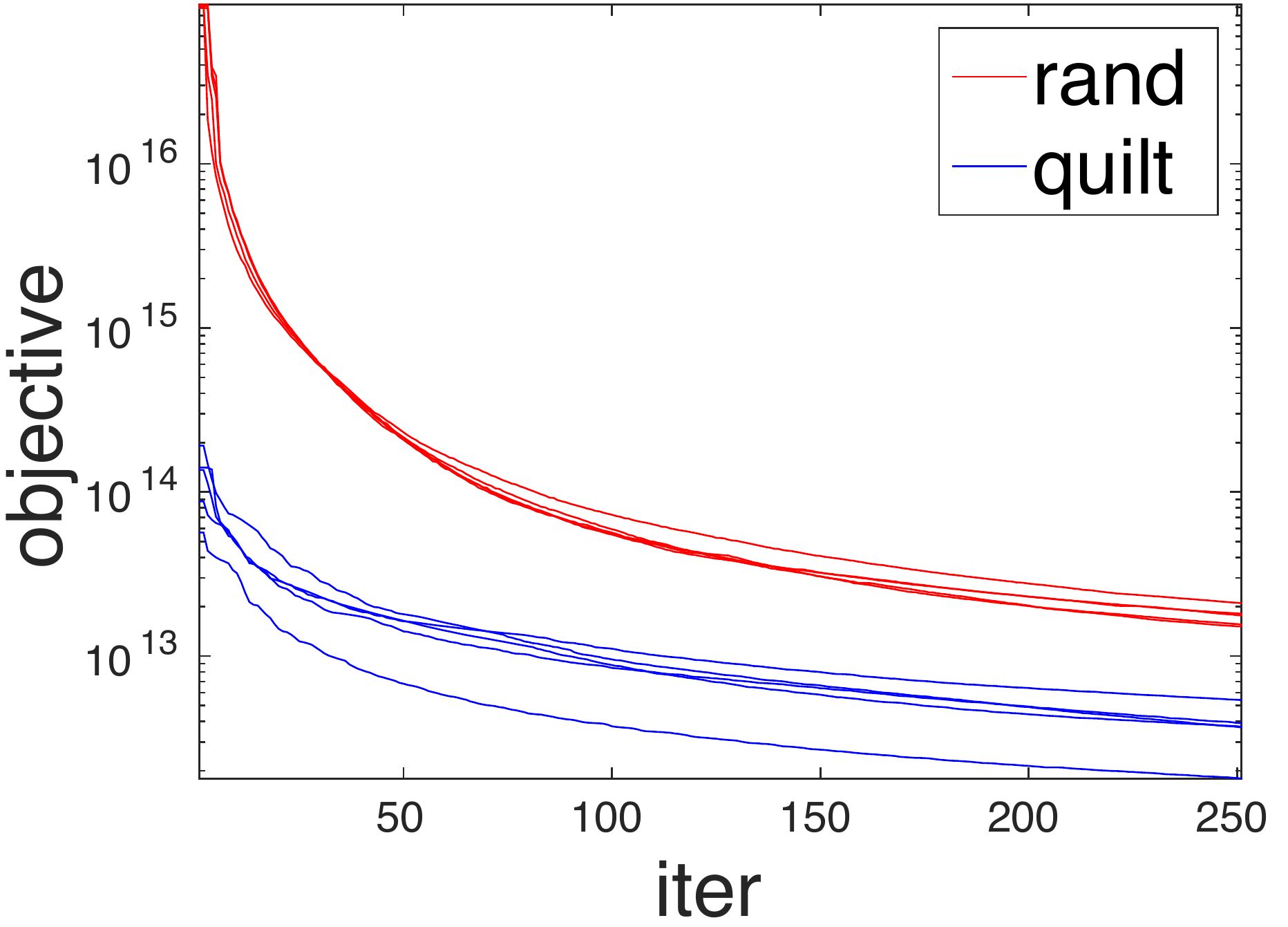} \\
\end{tabular}
\end{center}
\vspace{-0.1in}
\caption{\label{fig:tex-init} \textbf{Effect of initialization on texture synthesis.} Given an input image, the solution reached by the L-BFGS after 250 iterations starting from a random image: \emph{syn(rand)}, and image quilting: \emph{syn(quilt)}. The results using image quilting~\cite{Efros01} are shown as \emph{quilt}. On the right is the objective function for the optimization for 5 random initializations. Quilting-based initialization starts at a lower objective value and matches the final objective of the random initialization in far fewer iterations. Moreover, many artifacts of quilting are removed in the final solution (\eg, the top row). \emph{Best viewed with digital zoom}. Images are obtained from \url{http://www.textures.com}.}
\vspace{-0.1in}
\end{figure*}

\begin{figure}[h]
\begin{center}
\setlength{\tabcolsep}{1pt}
\renewcommand{\arraystretch}{0.6}
\begin{tabular}{ccc}
\puti{content}{\includegraphics[width=0.3\linewidth]{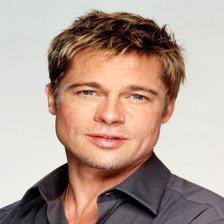}} & 
\puti{style}{\includegraphics[width=0.3\linewidth]{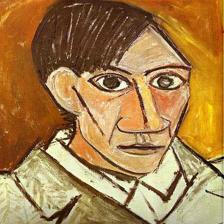}} & 
\puti{tranf(rand)}{\includegraphics[width=0.3\linewidth]{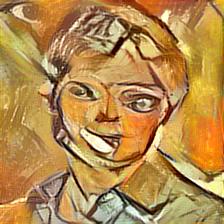}} \\
\puti{quilt}{\includegraphics[width=0.3\linewidth]{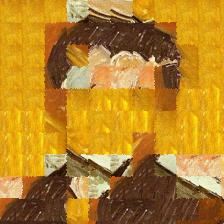}} & 
\puti{tranf(quilt)}{\includegraphics[width=0.3\linewidth]{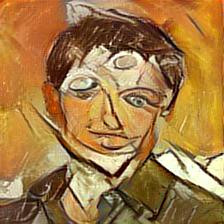}} & 
\includegraphics[width=0.3\linewidth]{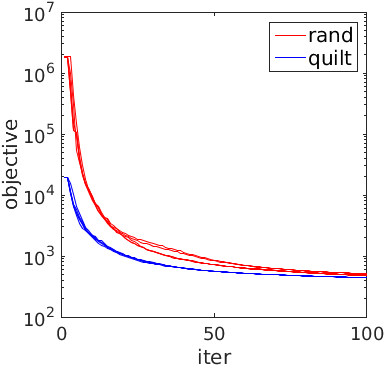} \\
\puti{content}{\includegraphics[width=0.3\linewidth]{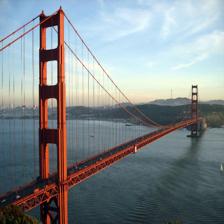}} & 
\puti{style}{\includegraphics[width=0.3\linewidth]{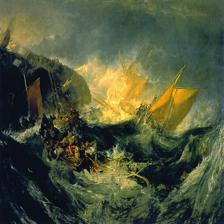}} & 
\puti{tranf(rand)}{\includegraphics[width=0.3\linewidth]{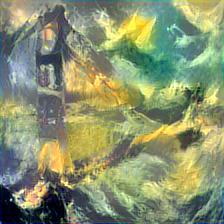}} \\
\puti{quilt}{\includegraphics[width=0.3\linewidth]{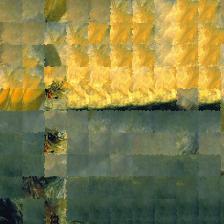}} & 
\puti{tranf(quilt)}{\includegraphics[width=0.3\linewidth]{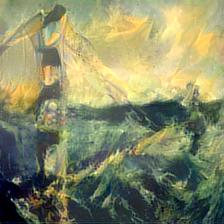}} & 
\includegraphics[width=0.3\linewidth]{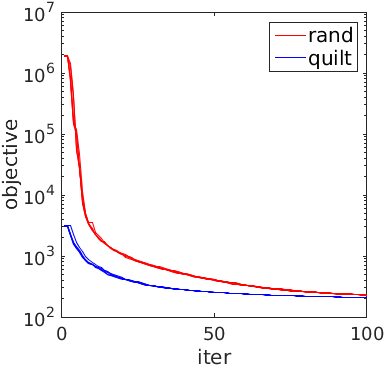} \\
\end{tabular}
\end{center}
\vspace{-0.1in}
\caption{\label{fig:style-init} \textbf{Effect of initialization on style transfer.} Given a content and a style image the style transfer reached using L-BFGS after 100 iterations starting from a random image: \emph{tranf(rand)}, and image quilting: \emph{tranf(quilt)}. The results using image quilting~\cite{Efros01} are shown as \emph{quilt}. On the bottom right is the objective function for the optimization for 5 random initializations.}
\vspace{-0.2in}
\end{figure}

\paragraph{Role of initialization on texture synthesis.} Although the same approach can be used for texture synthesis, it is not practical since it requires several hundreds of CNN evaluations, which takes several minutes on a high-end GPU. In comparison, non-parametric patch-based approaches such as \emph{image quilting}~\cite{Efros01} are orders of magnitude faster. Quilting introduces artifacts when adjacent patches do not align with each other. The original paper proposed an approach where a one-dimensional cut is found that minimizes artifacts. However, this can fail since local adjustments cannot remove large structural errors in the synthesis. We instead investigate the use of quilting to initialize the gradient-based synthesis approach. Fig.~\ref{fig:tex-init} shows the objective through iterations of L-BFGS starting from a random and quilting-based initialization. Quilting starts at a lower objective and reaches the final objective of the random initialization significantly faster. Moreover, the global adjustments of the image through gradient descent remove many artifacts that quilting introduces ({digitally zoom in to the \emph{onion image} to see this}). Fig.~\ref{fig:style-init} show the results using image quilting as initialization for style transfer~\cite{gatys2015neural}. Here two images are given as input, one for content measured as the \emph{conv4\_2} layer output, and one for style measured as the bilinear features. Similar to texture synthesis, the quilting-based initialization starts from lower objective value and the optimization converges faster. These experiments suggest that patch-based and parametric approaches for texture synthesis are complementary and can be combined effectively. 

\begin{figure}
\begin{center}
\setlength{\tabcolsep}{1.5pt}
\renewcommand{\arraystretch}{0.7}
\begin{tabular}{cccc}
\textit{relu2\_2} & + \textit{relu3\_3} & + \textit{relu4\_3} & + \textit{relu5\_3} \\
\includegraphics[width=0.24\linewidth]{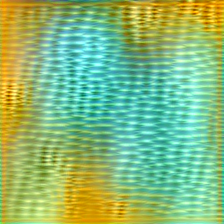} & 
\includegraphics[width=0.24\linewidth]{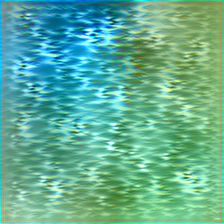} & 
\includegraphics[width=0.24\linewidth]{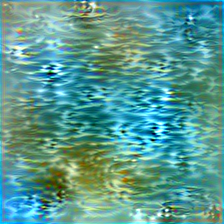} &
\puti{water}{\includegraphics[width=0.24\linewidth]{fig/inv/fmd/water.png}} \\
\includegraphics[width=0.24\linewidth]{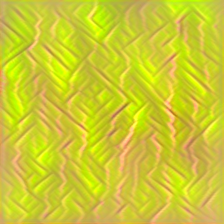} & 
\includegraphics[width=0.24\linewidth]{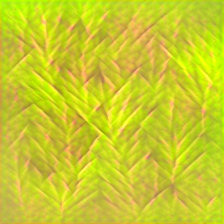} & 
\includegraphics[width=0.24\linewidth]{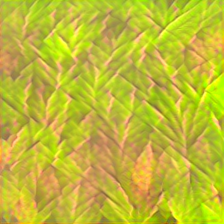} &
\puti{foliage}{\includegraphics[width=0.24\linewidth]{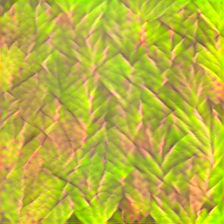}} \\
\includegraphics[width=0.24\linewidth]{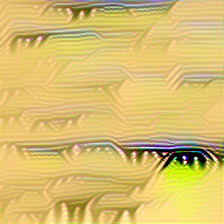} & 
\includegraphics[width=0.24\linewidth]{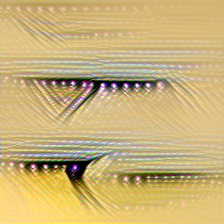} & 
\includegraphics[width=0.24\linewidth]{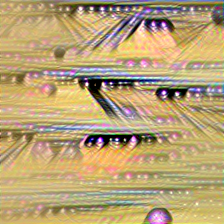} &
\puti{bowling}{\includegraphics[width=0.24\linewidth]{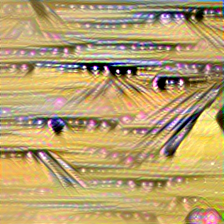}} \\
\end{tabular}
\end{center}
\vspace{-0.1in}
\caption{\label{fig:texture-layers} \textbf{Effect of layers on inversion.} Pre-images obtained by inverting class labels using different layers. The leftmost column shows inverses using predictions of \textit{relu2\_2} only. In the following columns we add layers \textit{relu3\_3}, \textit{relu4\_3}, and \textit{relu5\_3} one by one.}
\vspace{-0.2in}
\end{figure}

\vspace{-0.1in}

\begin{figure*}[th]
\begin{center}
\renewcommand{\arraystretch}{0.8}
\setlength{\tabcolsep}{1pt}
\begin{tabular}{cccccc}
\puti{braided}{\includegraphics[width=0.1615\linewidth]{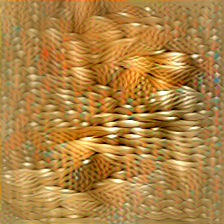}} & 
\puti{bubbly}{\includegraphics[width=0.1615\linewidth]{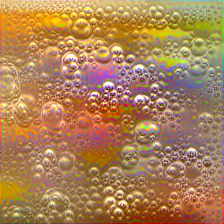}} & 
\puti{foliage}{\includegraphics[width=0.1615\linewidth]{fig/inv/fmd/foliage.png}} &
\puti{leather}{\includegraphics[width=0.1615\linewidth]{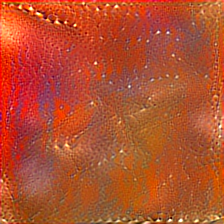}} &
\puti{bakery}{\includegraphics[width=0.1615\linewidth]{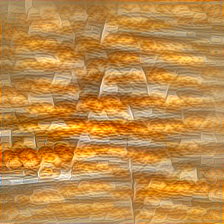}} &
\puti{bowling}{\includegraphics[width=0.1615\linewidth]{fig/inv/mit_indoor/bowling.png}} \\ 
\puti{cobwebbed}{\includegraphics[width=0.1615\linewidth]{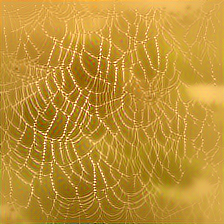}} & 
\puti{scaly}{\includegraphics[width=0.1615\linewidth]{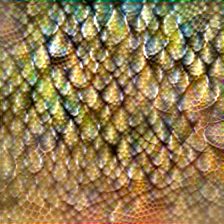}} & 
\puti{metal}{\includegraphics[width=0.1615\linewidth]{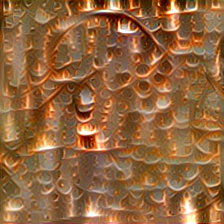}} &
\puti{stone}{\includegraphics[width=0.1615\linewidth]{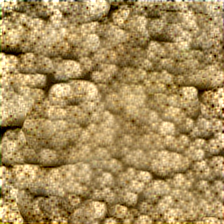}} &
\puti{classroom}{\includegraphics[width=0.1615\linewidth]{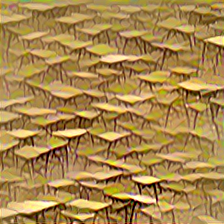}} &
\puti{closet}{\includegraphics[width=0.1615\linewidth]{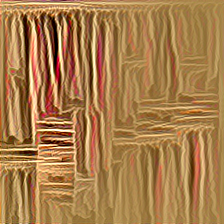}} \\ 
\end{tabular}
\end{center}
\vspace{-0.1in}
\caption{\label{fig:inverse-cont} \textbf{Examples of texture inverses (Fig.~\ref{fig:inverse} cont.)} Visualizing various categories by inverting the bilinear CNN model~\cite{lin2015bilinear} trained on DTD~\cite{cimpoi14describing}, FMD~\cite{sharan09material}, and MIT Indoor dataset~\cite{quattoni09recognizing} (two columns each from left to right). \emph{Best viewed in color and with zoom.}}
\vspace{-0.15in}
\end{figure*}

\paragraph{Visualizing texture categories.} We learn linear classifiers to predict categories using bilinear features from \emph{relu2\_2, relu3\_3, relu4\_3, relu5\_3} layers of the CNN on various datasets and visualize images that produce high prediction scores for each class. Fig.~\ref{fig:inverse} shows some example inverse images for various categories for the DTD, FMD and MIT indoor datasets. These images were obtained by setting $\beta=100$, $\gamma=1e-6$, and $\hat{C}$ to various class labels in Eqn.~\ref{eq:obj}. These images reveal how the model represents texture and scene categories. For instance, the \emph{dotted} category of DTD contains images of various colors and dot sizes and the inverse image is composed of multi-scale multi-colored dots. The inverse images of \emph{water} and \emph{wood} from FMD are highly representative of these categories. Note that these images cannot be obtained by simply averaging instances within a category which is likely to produce a blurry image. The orderless nature of the texture descriptor is essential to produce such sharp images. The inverse scene images from the MIT indoor dataset reveal key properties that the model learns -- a \emph{bookstore} is visualized as racks of books while a \emph{laundromat} has laundry machines at various scales and locations. In Fig.~\ref{fig:texture-layers} we visualize reconstructions by incrementally adding layers in the texture representation. Lower layers preserve color and small-scale structure and combining all the layers leads to better reconstructions. Even though the \emph{relu5\_3} layer provides the best recognition accuracy, simply using that layer did not produce good inverse images (not shown). Notably, color information is discarded in the upper layers. Fig.~\ref{fig:inverse-cont} shows visualizations of some other categories across datasets.

\begin{figure}
\begin{center}
\setlength{\tabcolsep}{1pt}
\renewcommand{\arraystretch}{0.6}
\begin{tabular}{ccc}
\puti{input}{\includegraphics[width=0.3\linewidth]{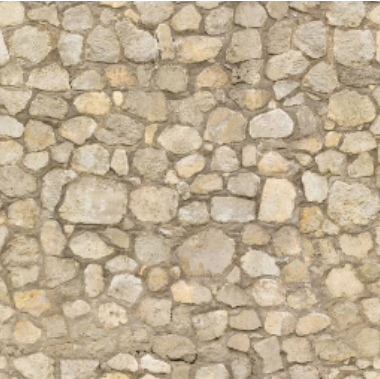}} & 
\puti{fibrous}{\includegraphics[width=0.3\linewidth]{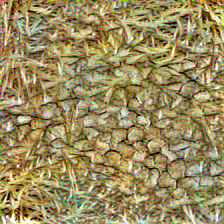}} & 
\puti{paisley}{\includegraphics[width=0.3\linewidth]{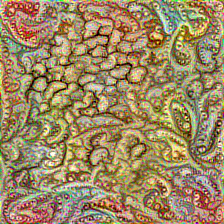}} \\
\puti{input}{\includegraphics[width=0.3\linewidth]{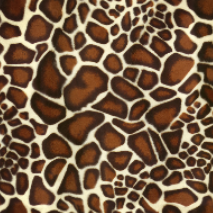}} & 
\puti{honeycombed}{\includegraphics[width=0.3\linewidth]{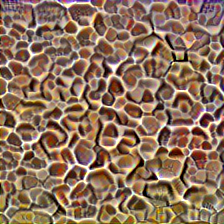}} & 
\puti{swirly}{\includegraphics[width=0.3\linewidth]{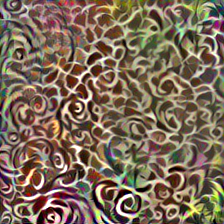}} \\
\puti{input}{\includegraphics[width=0.3\linewidth]{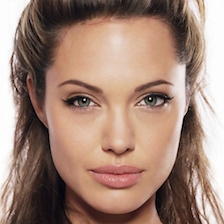}} & 
\puti{veined}{\includegraphics[width=0.3\linewidth]{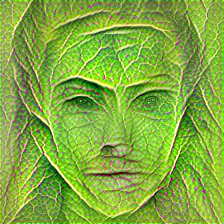}} & 
\puti{bumpy}{\includegraphics[width=0.3\linewidth]{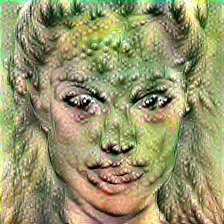}} \\
\puti{freckled}{\includegraphics[width=0.3\linewidth]{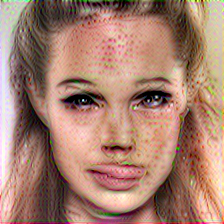}} & 
\puti{interlaced}{\includegraphics[width=0.3\linewidth]{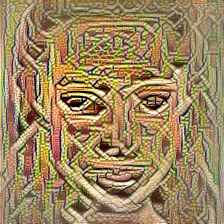}} & 
\puti{marbled}{\includegraphics[width=0.3\linewidth]{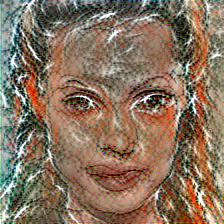}} \\
\end{tabular}
\end{center}
\vspace{-0.1in}
\caption{\label{fig:texture-edit} \textbf{Manipulating images with attributes.} Given an image we synthesize a new image that matches its texture (\emph{top two rows}) or its content (\emph{bottom two rows}) according to a given attribute (\emph{shown in the image}).}
\vspace{-0.25in}
\end{figure}

\begin{figure}[h]
\begin{center}
\setlength{\tabcolsep}{1pt}
\renewcommand{\arraystretch}{0.6}
\begin{tabular}{ccc}
\puti{chequered}{\includegraphics[width=0.3\linewidth]{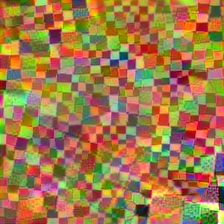}} &
\puti{$\beta_1/\beta_2 = 2.11$}{\includegraphics[width=0.3\linewidth]{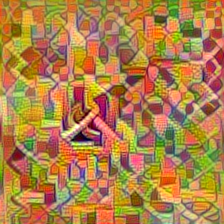}} & 
\puti{interlaced}{\includegraphics[width=0.3\linewidth]{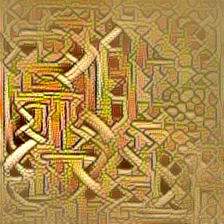}} \\
\puti{grid}{\includegraphics[width=0.3\linewidth]{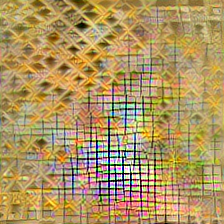}} &
\puti{$\beta_1/\beta_2 = 1.19$}{\includegraphics[width=0.3\linewidth]{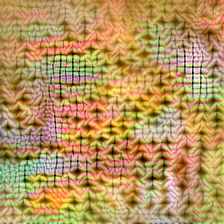}} &
\puti{knitted}{\includegraphics[width=0.3\linewidth]{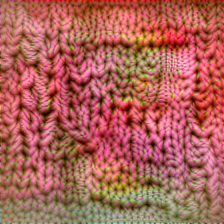}} \\
\puti{swirly}{\includegraphics[width=0.3\linewidth]{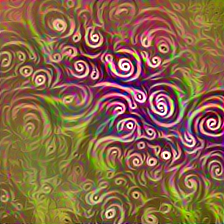}} & 
\puti{$\beta_1/\beta_2 = 0.75$ }{\includegraphics[width=0.3\linewidth]{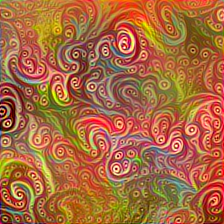}} & 
\puti{paisley}{\includegraphics[width=0.3\linewidth]{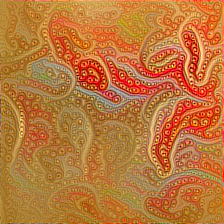}} \\
\end{tabular}
\end{center}
\vspace{-0.1in}
\caption{\label{fig:hybrid} \textbf{Hybrid textures} obtained by blending the texture on the left and right according to weights $\beta_1$ and $\beta_2$.}
\vspace{-0.1in}
\end{figure}

\section{Manipulating images with texture attributes}\label{sec:manipulation}
Our framework can be used to edit images with texture attributes. For instance, we can make a texture or the content of an image more honeycombed or swirly. Fig.~\ref{fig:texture-edit} shows some examples where we have modified images with various attributes. The top two rows of images were obtained by setting $\alpha_i=1~ \forall i$, $\beta=1000$ and $\gamma=1e-6$ and varying $\hat{C}$ to represent the target class. The bottom row is obtained by setting $\alpha_i=0~\forall i$, and using the \emph{relu4\_2} layer for content reconstruction with weight $\lambda=5e-8$. 

The difference between the two is that in the content reconstruction the overall structure of the image is preserved. The approach is similar to the neural style approach~\cite{gatys2015neural}, but instead of providing a style image we adjust the image with attributes. This leads to interesting results. For instance, when the face image is adjusted with the interlaced attribute (Fig.~\ref{fig:texture-edit} bottom row) the result matches the scale and orientation of the underlying image. No single image in the DTD dataset has all these variations but the categorical representation does. The approach can be used to modify an image with other high-level attributes such as artistic styles by learning style classifiers.

We can also blend texture categories using weights $\beta_j$ of the targets $\hat{C}_j$. Fig.~\ref{fig:hybrid} shows some examples. On the left is the first category, on the right is the second category, and in the middle is where a transition occurs (selected manually).

\section{Conclusion}\label{s:conclusion}

We present a systematic study of recent CNN-based texture representations by investigating their effectiveness on recognition tasks and studying their invariances by inverting them. The main conclusion is that translational invariance is a useful property not only for texture and scene recognition, but also for general object classification on the ImageNet dataset. The resulting models provide a rich parametric approach for texture synthesis and manipulation of content of images using texture attributes. The key challenge is that the approach is computationally expensive, and we present an initialization scheme based on image quilting that significantly speeds up the convergence and also removes many structural artifacts that quilting introduces. The complementary qualities of patch-based and gradient-based methods may be useful for other applications.

\vspace{0.1in}
\noindent
\textbf{Acknowledgement~~} The GPUs used in this research were generously donated by NVIDIA.

\FloatBarrier
\clearpage
{\small
\bibliographystyle{ieee}
\bibliography{bibliography/bcnn,bibliography/bibliography,bibliography/vgg_local,bibliography/vgg_other}
}

\end{document}